\documentclass[letterpaper, 10 pt, conference]{ieeeconf}  

\pdfoutput=1
\IEEEoverridecommandlockouts                              

\usepackage{graphicx}
\usepackage{subcaption}
\usepackage{algorithm}
\usepackage[noend]{algpseudocode}
\usepackage{amsmath} 
\usepackage{amssymb}  
\usepackage{myMacros}
\usepackage{soul}

\usepackage[dvipsnames]{xcolor}
\usepackage{comment}

\title{\LARGE \bf
Stability-Guaranteed Reinforcement Learning for Contact-rich Manipulation
}

\author{Shahbaz A. Khader$^{1,2}$, Hang Yin$^{1}$, Pietro Falco$^{2}$ and Danica Kragic$^{1}$%
\thanks{This work was supported by the Wallenberg AI, Autonomous Systems and Software Program (WASP) funded by the Knut and Alice Wallenberg Foundation.}
\thanks{$^{1}$Robotics, Perception, and Learning lab, Royal Institute of Technology, Stockholm, Sweden.
{\tt\small \{shahak, hyin, dani\}@kth.se}.}%
\thanks{$^{2}$ASEA Brown Boveri (ABB) Corporate Research, Sweden.
{\tt\small pietro.falco@se.abb.com}.}%
\thanks{Correspondence to {\tt\small shahak@kth.se}.}
}

\begin{document}

\maketitle
\thispagestyle{empty}
\pagestyle{empty}

\begin{abstract}
Reinforcement learning (RL) has had its fair share of success in contact-rich manipulation tasks but it still lags behind in benefiting from advances in robot control theory such as impedance control and stability guarantees. Recently, the concept of variable impedance control (VIC) was adopted into RL with encouraging results. However, the more important issue of stability remains unaddressed. To clarify the challenge in stable RL, we introduce the term \textit{all-the-time-stability} that unambiguously means that every possible rollout will be stability certified. Our contribution is a model-free RL method that not only adopts VIC but also achieves \textit{all-the-time-stability}. Building on a recently proposed stable VIC controller as the policy parameterization, we introduce a novel policy search algorithm that is inspired by Cross-Entropy Method and inherently guarantees stability. Our experimental studies confirm the feasibility and usefulness of stability guarantee and also features, to the best of our knowledge, the first successful application of RL with \textit{all-the-time-stability} on the benchmark problem of peg-in-hole.
\end{abstract}

\section{INTRODUCTION}
\label{sct:intro}

In contact-rich manipulation, modeling and control of contacts are necessary for successful execution of the task. Traditional planning and control methods that specialize in free motion and obstacle avoidance do not address contact-rich manipulation adequately.  Planning and control of contact-rich tasks is exceedingly difficult, especially when precise knowledge of the geometry and location of the manipulator and its surroundings is not available. The control of manipulator-environment interaction under the presence of uncertainties is generally studied as \textit{interaction control}~\cite{chiaverini1999survey}, but the terms \textit{compliant manipulation}~\cite{kronander2015control} and more recently \textit{contact-rich manipulation}~\cite{martin2019iros,lee2019making,levine2015learning} have also been used.  While seeking an appropriate control solution, an important property to be satisfied is stability, without which widespread adoption of robots cannot be realized.

Most \textit{interaction control} methods assume the availability of a nominal reference trajectory, which again presupposes some degree of knowledge of the task geometry. Reinforcement learning (RL) has emerged as a promising paradigm that can alleviate this concern. Many recent works have applied deep RL methods for learning contact-rich manipulation policies that directly outputs torques or forces~\cite{levine2015learning,thomas2018learning,tamar2017learning}. Although valid in principle, these methods do not benefit from the rich theory of interaction control and more importantly do not guarantee stability. In an encouraging trend, a number of recent works have adopted variable impedance control (VIC), a well-known interaction control concept, into an RL policy and showed its usefulness~\cite{martin2019iros,buchli2011learning,rey2018learning}. Unfortunately, none of the above mentioned methods guarantee stability. To clarify the scope of the stability we are interested in, we define the term \textit{all-the-time-stability} as guaranteed stability for all possible rollouts in the RL process. Our notion of stability is control theoretic (Lyapunov sense) and may not be confused with the requirement of smooth chatter-free contact, common in some industrial applications, that is also sometimes called "contact stability" \cite{salehian2018dynamical}. 
\begin{figure}[t]
    \centering
    \includegraphics[width=.48\textwidth]{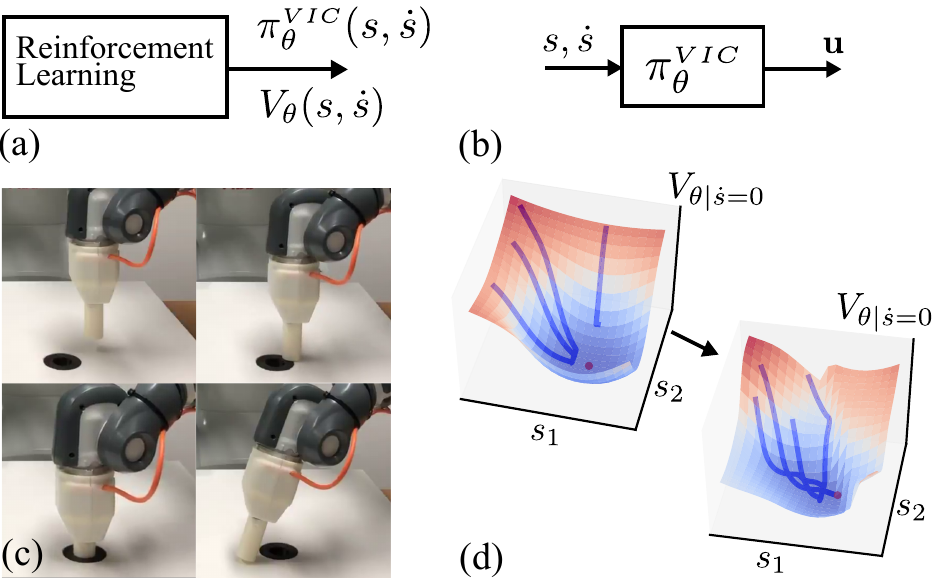}
    \caption{\textbf{(a)} RL optimizes both the policy ($\pi_{\theta}$) and Lyapunov function ($V_{\theta}$) \textbf{(b)} The VIC policy maps states (position, velocity) to action (force/torque). \textbf{(c)} Examples of contact states in peg-in-hole \textbf{(d)} RL reshapes untrained $V_{\theta}$ (illustrative) that cannot guide trajectories to the goal (red point) to one that can while maintaining \textit{all-the-time-stability}. 
    }  
    \label{fig:front_page}
\end{figure}

We propose a model-free RL algorithm with \textit{all-the-time-stability} that is well suited but not limited to contact-rich tasks. A model-free approach can avoid the complications of model learning \cite{khader2020data}. Our approach leverages a previously proposed framework for modeling stable VICs for discrete interaction motions~\cite{khansari2014modeling}. The main technical contribution is the introduction of a novel Evolution Strategy (ES) policy search method--inspired by Cross-Entropy Method (CEM)~\cite{cem-rubinstein-2003}--in which the sampling distribution is so constructed that it can only generate stable parameter samples. A novel update law is derived for iteratively updating the sampling distribution. Our experimental results show that not only \textit{all-the-time-stability} is possible without sacrificing sample efficiency, but it also confirms its usefulness. We demonstrate, to the best of our knowledge, the first successful reinforcement learning of the classical benchmark problem of peg-in-hole ~\cite{vgullapalli1992learning,lee2019making} with \textit{all-the-time-stability}.

\section{RELATED WORKS}
\label{sct:relatedworks}
This section is organized according to different topics:
\subsubsection{RL for Contact-rich Manipulation}
Learning artificial neural networks (ANN) policies for contact-rich manipulation has a long history~\cite{vgullapalli1992learning,nuttin1997learning}, but these early works adopted an admittance control approach that is more suited to non-rigid interactions~\cite{ott2010unified}. Recent examples are~\cite{levine2015learning,thomas2018learning,lee2019making,johannink2019residual}, where optimal control~\cite{levine2015learning}, motion planning~\cite{thomas2018learning}, multimodal feature learning~\cite{lee2019making} and residual learning \cite{johannink2019residual} were leveraged for making the problem tractable. Other approaches are model predictive control (MPC) with learned dynamics~\cite{lenz2015deepmpc,tamar2017learning} and learning reference force profile ~\cite{luo2019reinforcement}. None of these methods guarantee control stability during learning or for the final policy.

\subsubsection{Learning Variable Impedance Policies}\label{sec:lr_learning_VIC}
Learning a VIC policy can be done either through learning from demonstration (LfD)~\cite{Calinon10IROS,kronander2013learning,Khansari-Zadeh2017}, where human demonstration data is used to optimize policy parameters, or RL where a cost-driven autonomous trial-and-error process leads to a policy. A policy with VIC structure incorporates two levels of control loops: a trajectory generation loop and an impedance control loop. Guarantees on stability can be obtained only if both loops are considered in a unified way~\cite{khansari2014modeling}. An example of LfD with stability guarantee is \cite{Khansari-Zadeh2017}. 

In this paper we adopt the paradigm of RL which has the benefit of alleviating the demand of human demonstrations. Time-dependent policies, without any stability guarantees, were learned in \cite{buchli2011learning} and \cite{caldwell2010emvic}. In \cite{buchli2011learning} a Dynamical Movement Primitive (DMP) that encodes both reference and joint impedance trajectories is used, whereas in \cite{caldwell2010emvic}, a policy parameterized as a mixture of proportional-derivative systems is adopted. State-dependent policies were learned in ~\cite{rey2018learning,martin2019iros}. In~\cite{rey2018learning}, both the reference and stiffness trajectories are predicted from a policy parameterized by Gaussian Mixutre Regression (GMR) model. Interestingly, stability was guaranteed for the trajectory generation loop, but the overall stability was unclear because of the lack of unified analysis of both loops. The ANN based method~\cite{martin2019iros} learns a state-dependent policy that outputs reference trajectory and impedance parameters but offers no stability guarantees.

\subsubsection{Stable Variable Impedance Controllers}
A recent work~\cite{kronander2016stability} performed stability analysis for already existing trajectory and impedance profiles, but does not propose any policy structure for RL. In another interesting work~\cite{ferraguti2013tank}, the stability issue of VIC was tackled by a passivity-based approach, but it too assumed a reference trajectory to be given. Khanzari et al.~\cite{khansari2014modeling} proposed a trajectory-independent modeling framework i-MOGIC for discrete motions, featuring VIC and stability guarantees. Our method adopts i-MOGIC as the policy parameterization and makes progress to form a model-free RL algorithm with \textit{all-the-time-stability}. 

\subsubsection{Cross-Entropy Method for RL}
The Cross-Entropy method \cite{cem-rubinstein-2003}, a general purpose sampling-based optimization method, has been previously used for policy search both in an unconstrained \cite{mannor2003cross} and constrained \cite{wen2018constrained} setting. Our approach can also be seen as a constrained policy search, but unlike \cite{wen2018constrained}, the constraint (symmetric positive definiteness) is automatically guaranteed by the choice of the Wishart sampling distribution. Furthermore, unlike most cases where a Gaussian sampling distribution allows an analytical maximum likelihood estimation (MLE) based update, our special sampling distribution that consists of both Gaussian and Wishart factors necessitates a novel approach. Our policy search method may be seen as a CEM-inspired Evolution Strategy rather than a faithful implementation of CEM.

\subsubsection{Relation to Safe RL}
Stability has been considered as a means for safe RL in \cite{berkenkamp2017safe,controlbarrier_aaai2019,variancereduct_icml2019}.
The method in \cite{berkenkamp2017safe} depends on smoothness properties of the learned Gaussian process (GP) dynamics model, something that is inappropriate for contact. \cite{controlbarrier_aaai2019} also used GP and is further limited to learning only the unactuated part of the dynamics, while the remaining is assumed to be a known control-affine model. Another example \cite{variancereduct_icml2019} assumes the availability of a stabilizing prior controller, which, along with the magnitude of the disturbance in the assumed nominal dynamics model, determines the region of guaranteed stability. In our case, such limitations do not exist since no model is learned; instead, our approach assumes a fully known model of the manipulator dynamics against which stability is derived. This is not a limitation because such models are easily available. Unknown interaction dynamics does not undermine stability as long as the interaction is passive (see Sec. \ref{sec:bckgrnd_imogic}).

\section{BACKGROUND AND PRELIMINARIES}
\label{sct:bckgrnd}
\subsection{Reinforcement Learning}\label{sec:bckgrnd_rl}
A Markov decision process is defined by a set of states $\mathcal{X}$, a set of actions $\mathcal{U}$, a reward function $r(\vxi,\vui$), an initial state distribution $p(\vxi[0])$, a time horizon $T$, and dynamics $\transprob$. In policy search RL, the goal is to obtain the optimal stochastic policy $\pi_{\theta}(\vui|\vxi)$ by maximizing the expected trajectory reward $\mathbb{E}_{\tau}[\sum_{t=0}^T r(\vxi, \vui)]$, where $\tau=(\vxi[0],\vui[0],...,\vxi[T])$ represents a sample from a distribution of trajectories $p(\tau)$ that is induced by $p(\vxi[0])$, $\pi_{\theta}(\vxi)$ and $\transprob$ over the time period $T$. This is usually done by a gradient-based ($\nabla_{\theta}J(\theta)$) search for the optimum value $\theta^*$ of the parameterized policy $\pi_{\theta}$:
\begin{equation}\label{eqn:rl_opt}
    \theta^* = \underset{\theta}{\operatorname{argmax}} J(\theta) =  \underset{\theta}{\operatorname{argmax}}\mathbb{E}_{\tau}[\sum_{t=0}^Tr(\vxi, \vui)],
\end{equation}

In this paper, we consider a deterministic policy ($\vui=\pi_{\theta}(\vxi)$) and deterministic dynamics ($\vxi[t+1]=f(\vxi,\vui)$). Nevertheless, the general form of the problem in (\ref{eqn:rl_opt}) remains unchanged since a distribution $p(\tau)$ can still be induced by $p(\theta)$ and $p(\vxi[0])$. $p(\theta)$ arises in the context of parameter space exploration \cite{Plappert18Parameter}, which is less common than action space exploration. Parameter space exploration allows exploration with deterministic policies.

\subsection{Stability in RL}\label{sec:bckgrnd_stab}
Lyapunov stability analysis deals with the study of time evolution of non-linear control systems such as robotic manipulators \cite{slotine1991applied}. An equilibrium point of the system is stable if the state trajectories that start close enough are bounded around it and asymptotically stable if the trajectories converge to it. Such properties are necessary to ensure that the system does not fail catastrophically in a wide range of circumstances. Stability is guaranteed by proving the existence of a certain Lyapunov function $V(\vxi)$--a radially unbounded, positive definite and scalar function of the state $\vxi$--that has the properties $V(\vec{x}^*)=0$ and $\dot{V}(\vec{x})<0\ \forall \vec{x}\ne\vec{x}^*$ and $\dot{V}(\vec{x}^*)=0$, where $\vec{x}^*$ is the equilibrium point. 
 
In an RL context, stability corresponds to a guarantee that any rollout is bounded in state space and tends to the goal position demanded by the task. Ideally, the policy should ensure the goal position as the unique equilibrium point, in which case we can refer to stability of the system and not just an equilibrium point.

\subsection{Cross Entropy Methods}
\label{scn:bkgrnd:cem}
Cross Entropy Methods (CEM)~\cite{cem-rubinstein-2003} conducts sampling-based optimization to solve problems like~\eqref{eqn:rl_opt}. The optimization relies on a sampling distribution $q(\theta | \mathbf{\Phi})$ to generate the samples $\{\theta_n \}_{n=1}^{N_s}$. The performance of each $\theta_n$ is evaluated according to $J(\theta_n)$ and used to compute updates for the distribution parameter $\mathbf{\Phi}$. Formally, it iteratively solves:
\begin{equation}\label{eqn:cem}
    \mathbf{\Phi}^{i+1} = \underset{\mathbf{\Phi}}{\operatorname{argmax}}\sum\limits_{n}\mathbb{I}(J(\theta_n^i)) \log q(\theta_n^i|\mathbf{\Phi})   \quad     \theta_n^i \sim q(\theta|\mathbf{\Phi}^i)
\end{equation}
where $\mathbb{I}(\cdot)$ denotes an indicator function that selects only the best $N_e(<N_s)$ samples or \textit{elites} based on individual performance $J(\theta_n)$. The computation of $\mathbf{\Phi}^{i+1}$ from the samples $\{\theta_n^i \}_{n=1}^{N_e}$ is done by MLE. Very often $q(\theta | \mathbf{\Phi})$ is modeled as Gaussian, for which an analytical solution exists~\cite{cem-rubinstein-2003}. As the iteration index $i$ grows, $q(\theta|\mathbf{\Phi}^i)$ is encouraged to converge to a narrow solution distribution with high performance, thereby approximately solving~\eqref{eqn:rl_opt}.

\subsection{i-MOGIC}\label{sec:bckgrnd_imogic}
Integrated MOtion Generator and Impedance Controller (i-MOGIC) was proposed as a modeling framework for discrete interaction motions~\cite{khansari2014modeling}. The controller has the form of a weighted mixture of several spring-damper components:
\begin{equation}\label{eqn:imogic}
\begin{split}
    \vu = -\boldsymbol{S}^0s - \boldsymbol{D}^0\Dot{s} & - \sum\limits_{k=1}^K w^k(s) [\boldsymbol{S}^k(s - s^k) +  \boldsymbol{D}^k\Dot{s}]   \\
\end{split}
\end{equation}
The stiffness and damping matrices are denoted by $\boldsymbol{S}$ and $\boldsymbol{D}$, respectively. The superscript $0$ is used to indicate the base spring-damper and $k$ for a mixture component. $s$ and $\Dot{s}$ denote position and velocity of the manipulator. Note that $s$ is defined relative to the goal position, implying that the global attractor point is the origin. The attractor points for the remaining components are denoted by $s^k$.
The mixing weight $w^k(s)$ is a function of $s$ and parameterized by $\boldsymbol{S}^k,s^k,l^k$, where $l^k$ is a scalar quantity.

Equation (\ref{eqn:imogic}) can be seen as a VIC policy \cite{khansari2014modeling} of the form:
\begin{equation}\label{eqn:vic_policy}
    \vec{u}=\pi_{\theta}(\vec{x})=-\Bar{\boldsymbol{S}}(\vec{x})(s-\Bar{s}(\vec{x}))-\Bar{\boldsymbol{D}}(\vec{x})\dot{s},
\end{equation}
where $\vec{x}=[s^T, \dot{s}^T]^T$, $\vec{u}$ is force/torque and $\Bar{s}$, $\Bar{\boldsymbol{S}}$ and $\Bar{\boldsymbol{D}}$ are state-dependent position reference, stiffness matrix and damping matrix, respectively. The parameter set of i-MOGIC is given by: 
\begin{equation}\label{eqn:param_set}
\theta = \{\boldsymbol{S}^0, \boldsymbol{D}^0, \boldsymbol{S}^k, \boldsymbol{D}^k, s^k, l^k\} \text{ for } k=1, ..., K.    
\end{equation}
Khansari et al.~\cite{khansari2014modeling} showed that, with gravity compensated, (\ref{eqn:imogic}) is globally asymptotically stable (GAS) at the origin if:
\begin{equation}\label{eqn:imogicstablecon}
    \begin{split}
        \boldsymbol{S}^0 = (\boldsymbol{S}^0)^T \succ 0 \quad \boldsymbol{D}^0 \succ 0 \\
        \boldsymbol{S}^k = (\boldsymbol{S}^k)^T \succeq 0 \quad \boldsymbol{D}^k \succeq 0 \quad l^k > 0 \quad \forall k = 1, ..., K 
    \end{split}
\end{equation}
GAS and \textit{passivity} is proven for free motion using only the manipulator dynamics while considering any interaction forces as persistent disturbance. When the interaction is with a \textit{passive} environment, asymptotic stability may be lost but stability is retained \cite{khansari2014modeling}. A simple example is when an obstacle is preventing the manipulator from moving towards the goal; here, the contact forces constitute the persistent disturbance. A \textit{passive} environment corresponds to the common case where objects in the environment are not actuated. Note that some terms in the original i-MOGIC are omitted in (\ref{eqn:imogic}) since they are set to zero.

\section{APPROACH}\label{sct:methods}
\begin{figure}[t]
    \centering
    \includegraphics[width=.5\textwidth,trim=1cm 2.5cm 0cm 2cm, clip]{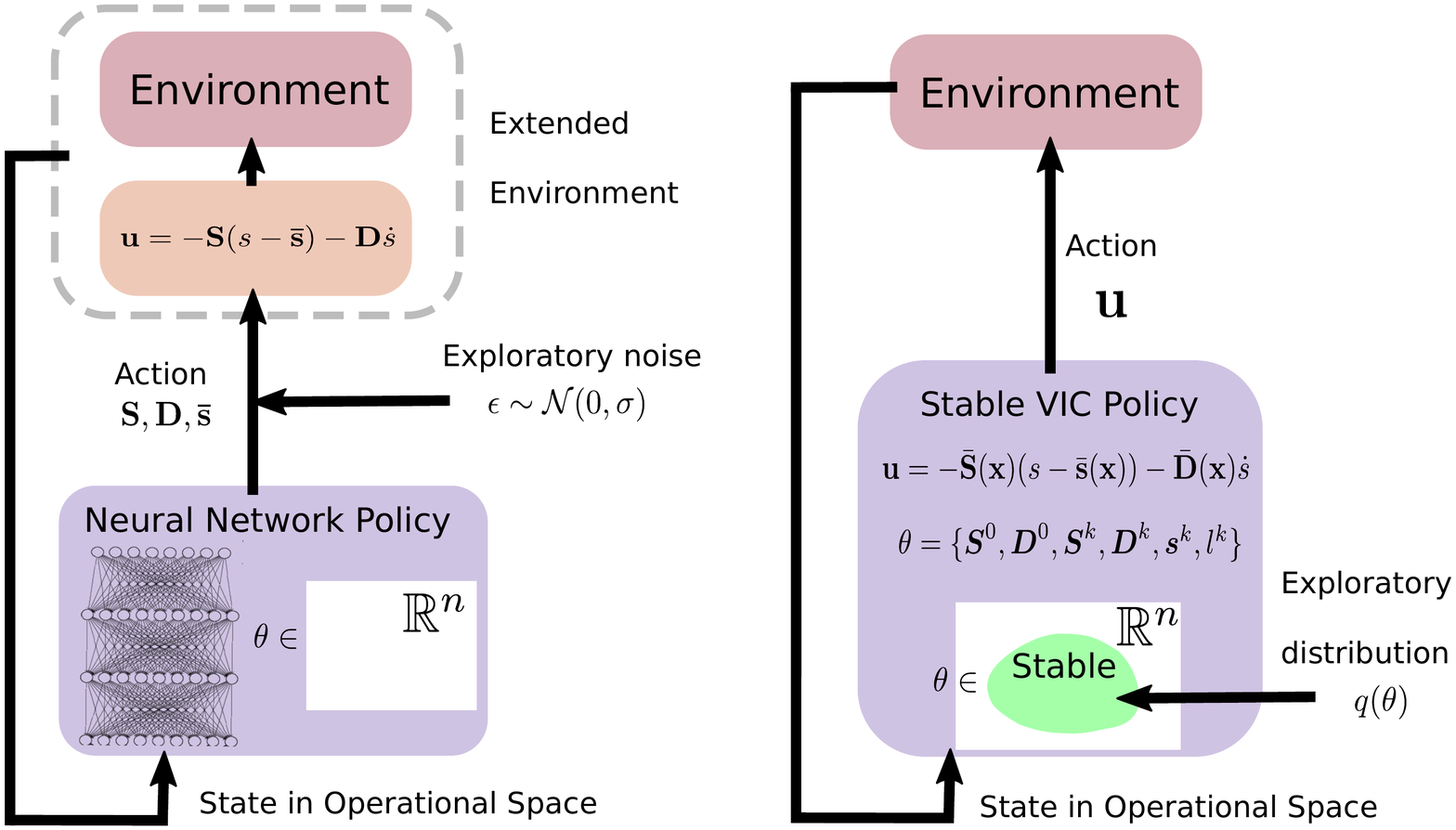}
    \caption{RL with VIC policy structure. (\textbf{Left}): stability-unaware neural network policies outputting unconstrained impedance gains (\textbf{Right}): \textit{all-the-time-stability} policy search with interpretable and constrained parameters. 
    }  
    \label{fig:comp_vicrl}
\end{figure}
As we have seen, i-MOGIC is a parameterized policy with a VIC structure and stability guarantees. To attain our goal of an RL algorithm with \textit{all-the-time-stability}, we need a policy search algorithm that also features stable exploration and convergence guarantees. Our main contribution is a model-free RL algorithm that meets these requirements. We adopt i-MOGIC policy parameterization and propose a CEM-like policy search algorithm. i-MOGIC being a deterministic policy and its stability determined solely by its parameter values, the parameter space exploration strategy of CEM is ideal. Figure \ref{fig:comp_vicrl} provides further perspective by juxtaposing our solution with a deep RL approach with VIC structure and the more common action space exploration \cite{martin2019iros}.

In our novel CEM-like algorithm, the constraints in (\ref{eqn:imogicstablecon}) are guaranteed by designing a sampling distribution that makes it impossible to sample an unstable parameter set. A feasible parameter set of i-MOGIC is a mixed set of real-valued vectors, positive scalars, and matrices with symmetry and positive (semi)definiteness. Considering positive numbers as a special case of symmetric positive definite (SPD) matrices and enforcing SPD for all the matrices in (\ref{eqn:param_set}), all parameters except $s^k$ can be modeled by a distribution of SPD to guarantee constraint satisfaction. We focus on this aspect first and then develop the complete solution subsequently. Note that enforcing SPD for all matrices in (\ref{eqn:param_set}) implies that our approach is slightly more conservative than it is required.

\subsection{Optimizing Positive-definite Matrix Parameters}\label{sec:wishart_update}
\begin{figure}[t]
    \centering
    \includegraphics[width=.49\textwidth]{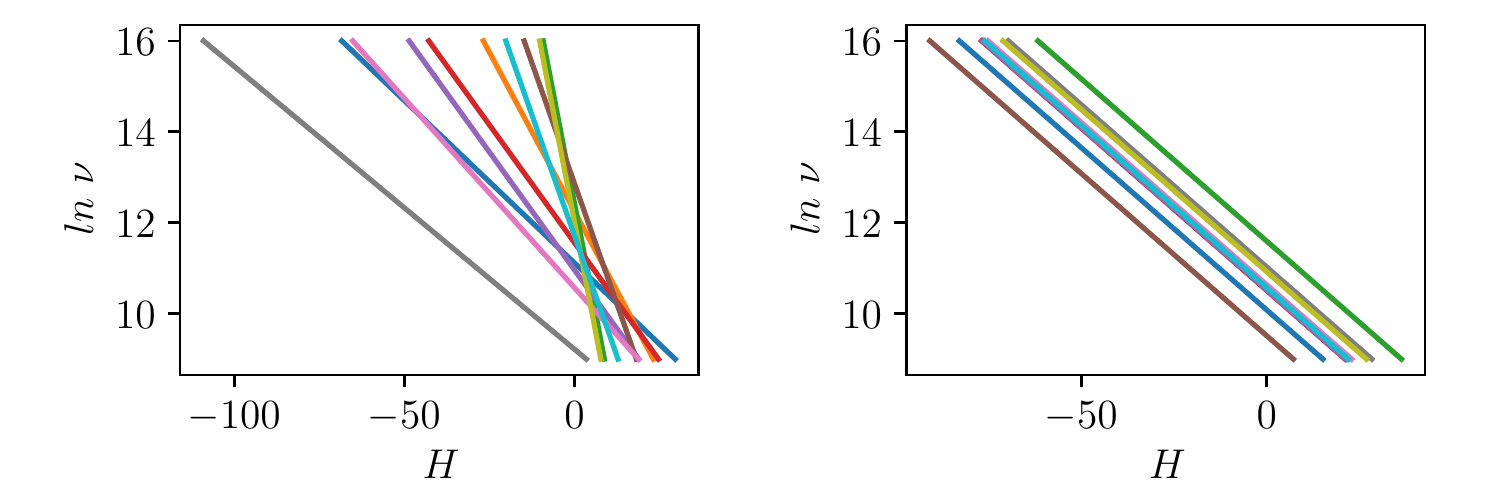}
    \caption{The (apparent) linear relationship between $ln~\nu$ and entropy $H$ of a Wishart distribution. $\mathbb{E}[S]=\nu W$ is fixed and $\nu>D+1$. Each plot represents a different distribution. \textbf{Left}: random dimensions ($D$ between 2 and 10) and random $W$ parameters. \textbf{Right}: $D=7$ and and random $W$. }  
    \label{fig:wishart_linear}
\end{figure}
We model the sampling distribution for an SPD matrix with the Wishart distribution, which is fully defined by two parameters $\nu$ and $W$ and denoted by $S\sim \mathcal{W}_D(S|W,\nu)$. $S\in\mathbb{R}^{D\times D}$ and $W\in\mathbb{R}^{D\times D}$ are SPD and $\nu>D-1$. The expected value is given by $\mathbb{E}[S] = \nu W$. The variance of a random matrix is not easy to define but is controlled by $\nu$. MLE-based update of Wishart distribution is possible if we adopt a numerical optimization approach for (\ref{eqn:cem}). Often, it is desirable to avoid such an approach because it may entail specification of parameter intervals, gradient computation or lack of convergence guarantees. Instead, we derive a simple update rule that, in spite of not targeting to solve (\ref{eqn:cem}), has the general property of continuously refining the sampling distribution towards an approximate solution. Although ours is no longer a faithful CEM method, it still conforms to the Evolution Strategy paradigm.

We propose to update the $W$ parameter of the Wishart distribution by the empirical average of the elite samples $\{S_m\}_{m=1}^{N_e}$. This is valid since scaling and addition preserves symmetry and positive definiteness. Our update rule for the $\nu$ parameter is based on two modeling assumptions: $\Delta H \propto -\frac{(R_e - R_b)}{R_b}$ and $\Delta \ln~\nu \propto -\Delta H$, where $H$ is the entropy of the Wishart distribution; and $R_b$ and $R_e$ are the average rewards of all samples and elites, respectively. The first linear model is our design choice and is motivated by the intuition that there should be a reduction in entropy of the sampling distribution commensurate with a relative gain between $R_b$ and $R_e$. The second model actually matches closely with reality as evidenced by Fig. \ref{fig:wishart_linear}. After combining the linear models and rearranging terms we obtain our update law:
\begin{subequations}\label{eqn:update_law}
\begin{align}
    W^{i+1}=\frac{1}{N_e}\sum_{m=1}^{N_e}\Bar{S}_m^{i},\\
    \nu^{i+1} = \nu^{i}\text{exp}\left(\gamma\beta \left(\frac{R_e^i - R_b^i}{R_b^i}\right)\right)
\end{align}
\end{subequations}
where $\Bar{S}_m^{i}$ is the elite sample scaled by $\frac{1}{\nu}$, $i$ is the iteration variable (see (\ref{eqn:cem})), and $\beta>0$ and $\gamma>0$ are the respective proportionality constants of the linear models. Since Fig. \ref{fig:wishart_linear} also indicates that $\gamma$ is dependent only on $D$, $\gamma$ can be estimated once $D$ is known (for example, $D=dim(s)$). The constant $\beta$, on the other hand, is a hyperparameter and its tuning corresponds to the trade-off between exploration and exploitation. Higher values can achieve updates that favour exploitation instead of exploration because the variance of Wishart distribution is a decreasing function of $\nu$.

We now focus on the convergence behavior of the proposed method. Notice that $R_e\ge R_b$ and therefore $\nu^{i+1}\ge \nu^{i}$. The boundary condition $R_e=R_b$ is satisfied only at convergence where all samples from the sampling distribution return identical rewards. In all other cases the distribution shrinks because of the decreasing nature of variance with increasing $\nu$. This guarantees eventual convergence. Indeed, the converged solution may be a local optimum, as it would be the case for any policy search method. It is worth noting that although our modeling assumption, $\ln~\nu \propto -H$, is most appropriate, any decreasing function $\nu$ of $H$ is sufficient.

\subsection{Stability-Guaranteed Policy Search}\label{sec:method}
\begin{algorithm}[t]
\caption{Policy search with stability guarantees}\label{alg:man_algo}
\begin{algorithmic}[1]
\State Initialize $i=0$;
\State $\boldsymbol{\Phi}^0=\{\boldsymbol{\Phi}_{\vec{s}}^0,\boldsymbol{\Phi}_{\boldsymbol{S}^0}^0,\boldsymbol{\Phi}_{\boldsymbol{D}^0}^0,\boldsymbol{\Phi}_{\boldsymbol{S}^k}^0,\boldsymbol{\Phi}_{\boldsymbol{D}^k}^0,\boldsymbol{\Phi}_{l^k}^0\}$ for $k=\{1,...,K\}$
\While{not converged}
\State Do rollouts with $\theta_n^i \sim q(\theta_n^i|\boldsymbol{\Phi}^i)$, $n=1...N_s$
\State Extract $N_e$ samples from $N_s$ based on performance measure $J(\theta_n^i)$ \Comment{Sec. \ref{scn:bkgrnd:cem}}
\State Compute $\boldsymbol{\Phi}_{\vec{s}}^{i+1}$ using MLE on $\{\vec{s}_m^i\}_{m=1}^{N_e}$
\State Compute $\boldsymbol{\Phi}_{\boldsymbol{M}}^{i+1}$ using (\ref{eqn:update_law}) on $\{\boldsymbol{M}_m^i\}_{m=1}^{N_e}$,
\State \qquad $\boldsymbol{M\in\{\boldsymbol{S}^0,\boldsymbol{D}^0,\boldsymbol{S}^k,\boldsymbol{D}^k,l^k\}}$ for $k=1,...,K$
\State Increment $i$
\EndWhile
\end{algorithmic}
\end{algorithm}

Our goal is to derive a model-free policy search method that guarantees the stability constraints of i-MOGIC. More specifically, we seek an algorithm that solves (\ref{eqn:rl_opt}) using the general strategy in (\ref{eqn:cem}), if possible, but subject to the stability constraints in (\ref{eqn:imogicstablecon}). The policy parameterization $\pi_\theta$ is (\ref{eqn:imogic}), with the parameter set $\theta$ in (\ref{eqn:param_set}). The only missing piece is the sampling distribution $q$ and the exact strategy for the iterative update of its parameter $\boldsymbol{\Phi}$.

We define a sampling distribution of the form:
\begin{align}\label{eqn:prop_dist}
\begin{split}
    q(\theta|\boldsymbol{\Phi}) = 
    q(\vec{s}|\boldsymbol{\Phi}_{\vec{s}})
    q(\boldsymbol{S}^0|\boldsymbol{\Phi}_{\boldsymbol{S}^0})
    q(\boldsymbol{D}^0|\boldsymbol{\Phi}_{\boldsymbol{D}^0})...\\
    \prod_{k=1}^K q(\boldsymbol{S}^k|\boldsymbol{\Phi}_{\boldsymbol{S}^k})
    q(\boldsymbol{D}^k|\boldsymbol{\Phi}_{\boldsymbol{D}^k})
    q(l^k|\boldsymbol{\Phi}_{l^k}),
\end{split}
\end{align}
where $q(\theta|\boldsymbol{\Phi})$ is the joint probability distribution of all the parameters in (\ref{eqn:param_set}). After introducing a notational simplification $\vec{s}=[{(s^1)}^T,...,{(s^K)}^T]^T$, $q(\vec{s}|\boldsymbol{\Phi}_{\vec{s}})$ is a multivariate Gaussian distribution with parameters $\boldsymbol{\Phi}_{\vec{s}}=\{\mu_{\vec{s}},\Sigma_{\vec{s}}\}$; and $q(\boldsymbol{M}|\boldsymbol{\Phi}_{\boldsymbol{M}})$ represents Wishart distributions with parameters $\boldsymbol{\Phi}_{\boldsymbol{M}}=\{W_M,\nu_{M}\}$ where $\boldsymbol{M}\in\{\boldsymbol{S}^0,\boldsymbol{D}^0,\boldsymbol{S}^k,\boldsymbol{D}^k,l^k\}$ for $k=1,...,K$.
While the matrix cases have dimension $D=\text{dim}(s)$, $q(l^k)$ is a one dimensional Wishart distribution of positive numbers $l^k\in\mathbb{R}^+$ (also equivalent to Gamma distribution). Since we have modeled the individual elements in $\theta$ as independent random variables, each of them can be maintained and updated independently. The update of the Gaussian $q(\vec{s}|\boldsymbol{\Phi}_{\vec{s}})$ is performed by the standard practice of MLE. For the remaining Wishart cases we employ our novel strategy in (\ref{eqn:update_law}). The complete algorithm is shown in Alg.~\ref{alg:man_algo}.

\section{EXPERIMENTAL RESULTS}
\label{sct:experiments}
Two experimental environments are defined: 2D block-insertion and robot peg-in-hole. VICES \cite{martin2019iros}, a deep RL method based on Proximal Policy Optimization that features VIC and no stability guarantee, was chosen as the baseline. We used [16,16] and [32,32] as the network structures for the first and second tasks respectively. The platforms consisted of the simulator MuJoCo and the robot YuMi, both updated at 100 Hz.

\subsection{Simulated 2D Block-insertion}\label{sec:result_blocks}
\begin{figure}[t]
    \centering
    \includegraphics[width=.43\textwidth]{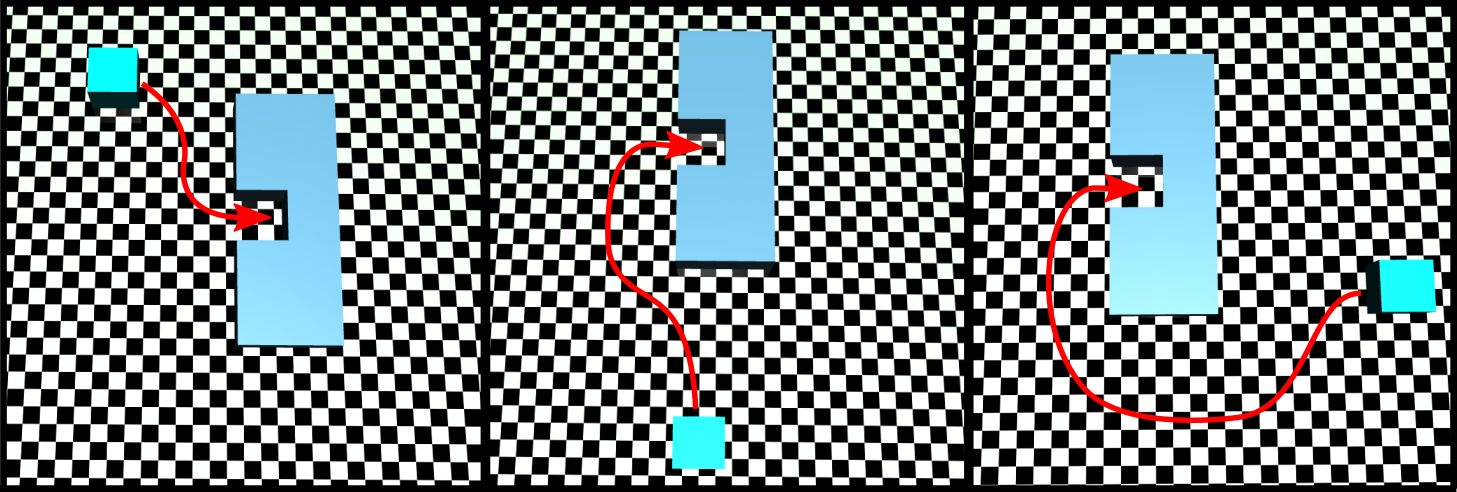}
    \caption{\textbf{2D block-insertion tasks.} From left to right: Task1 (insertion clearance: 0.5 mm, execution time: 1s), Task2 and Task3 (insertion clearance: 2 mm, execution time: 2s). Block size is 50$\times$50$\times$50 mm and weighs 2Kg. Rough illustrative paths are indicated by red arrows.} 
    \label{fig:blocks_exp_setup}
\end{figure}

In this experiment, three variations of the simulated 2D block-insertion tasks are set up by varying the initial position of the block and also the insertion clearance (Fig. \ref{fig:blocks_exp_setup}). The policy controls the block by exerting an orthogonal 2D force (rotation not allowed) with the goal of inserting it into the slot. The task executions are expected to generate contacts between the block and the environment.

\begin{figure}[t]
    \centering
    \includegraphics[width=.49\textwidth]{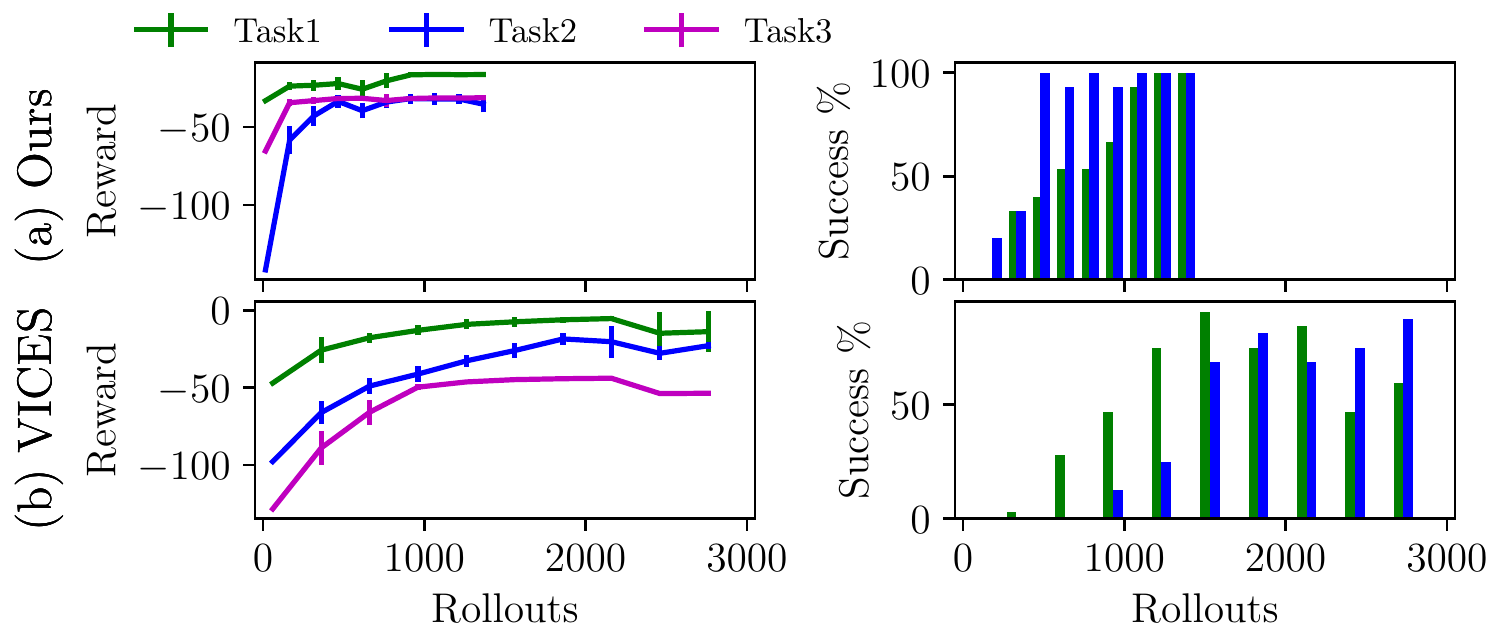}
    \caption{\textbf{RL of 2D block-insertion tasks.} \textbf{(a)} Our method for 50 iteration with $N_s=15$. \textbf{(b)} VICES for 50 iteration with $N_s=30$. Success rate is among all rollouts per iteration.}  
    \label{fig:blocks_all_tasks}
\end{figure}
\begin{figure}[t]
     \centering
     \begin{subfigure}[b]{0.49\textwidth}
         \centering
         \includegraphics[width=3.5in,height=1in]{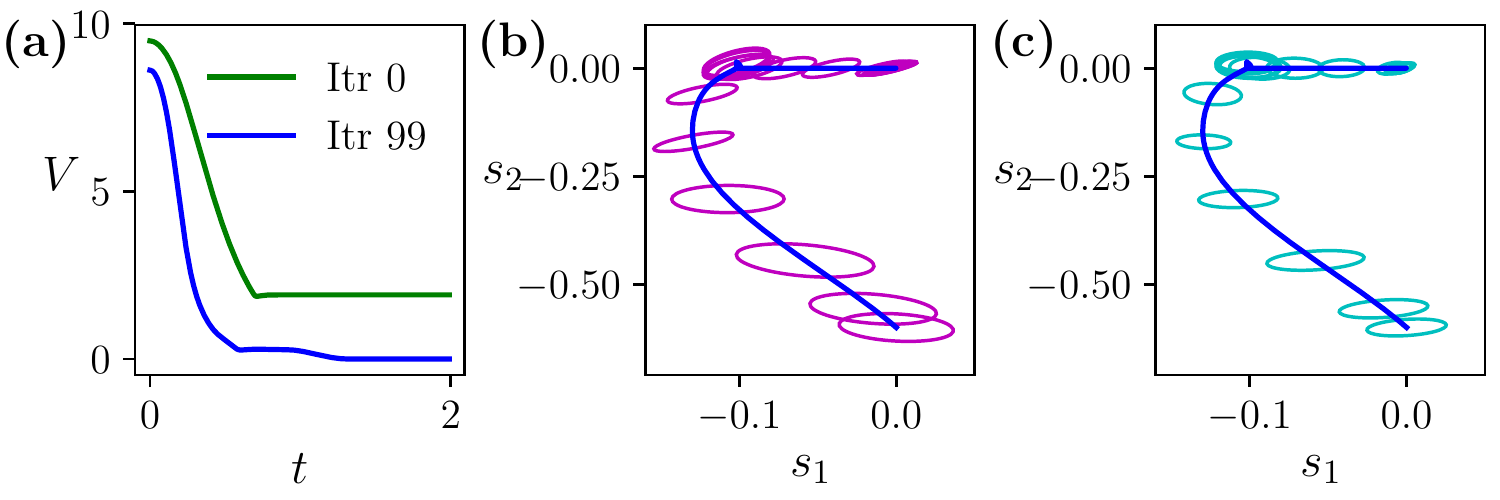}
     \end{subfigure} \\
     \begin{subfigure}[b]{0.49\textwidth}
         \centering
         \includegraphics[width=\textwidth]{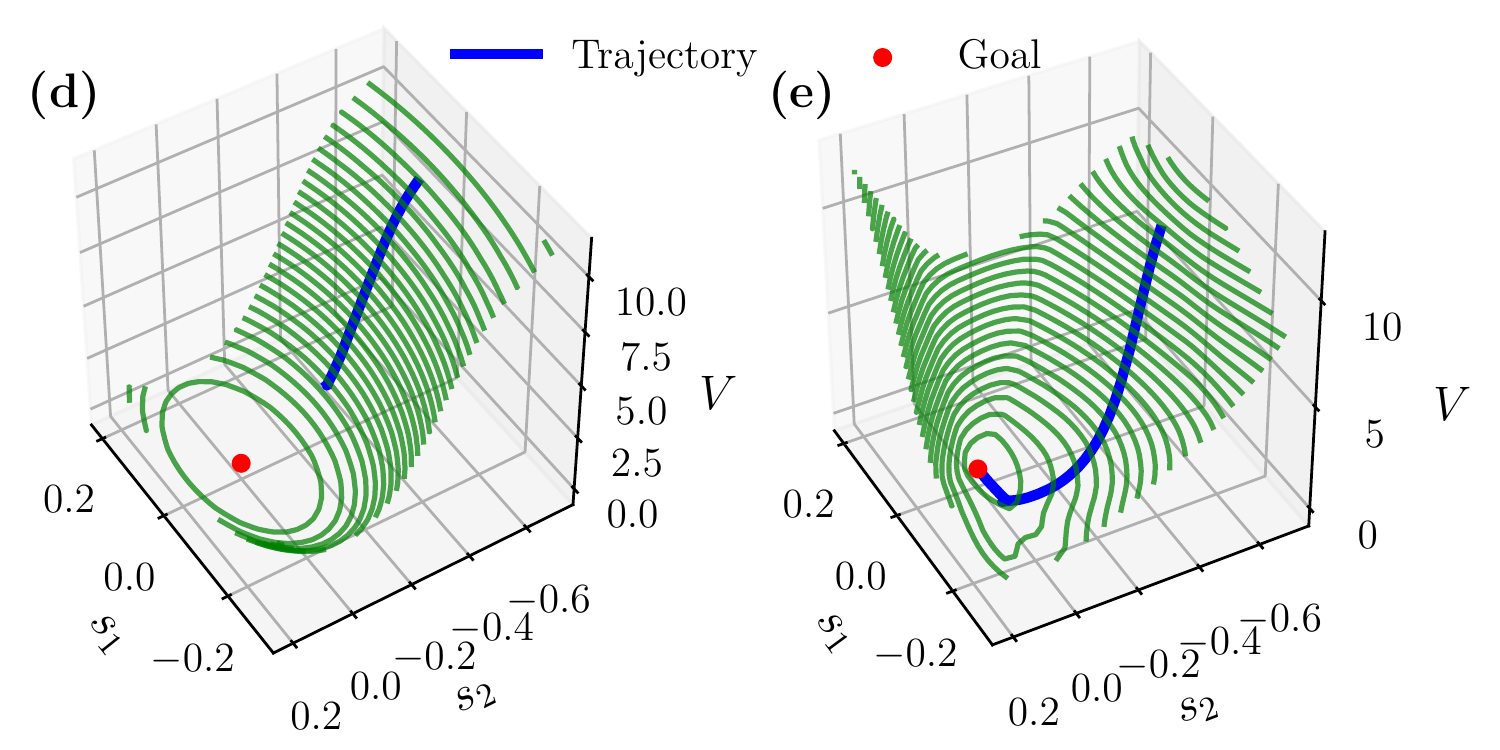}
     \end{subfigure}
     \caption{\textbf{VIC policy is learned while maintaining stability (Task2).} Two rollouts are considered: $\tau_0$ from iteration 0 (unsuccessful) and $\tau_{99}$ from iteration 99 (successful). \textbf{(a)} Lyapunov energy ($V(s,\dot{s})$) plots of the rollouts. Combined stiffness (\textbf{b}) and damping (\textbf{c}) matrices at regular intervals along $\tau_{99}$. (\textbf{d}-\textbf{e}) Contour plots of $V(s,\dot{s})_{|\dot{s}=0}$ for $\tau_0$ and $\tau_{99}$, respectively, with their trajectories overlaid on them.}
     \label{fig:blocks_task2_energy_comp}
\end{figure}

Other aspects include hyperparameter tuning, initialization and reward model. Through a process of grid search, we determined the number of spring-damper components ($K$) as $1$ and $8$ for Task1 and Task2 respectively. Task3 was not successful. Learning rate, $\beta=1$, was found to work well. We fixed $N_s=15$ and $N_e=3$ for all our experiments. This choice is motivated by the real-world sample complexity of RL for our final experiment, where an excess of $15$ rollouts per iteration may be undesirable. We initialized $q(\theta|\Phi)$ in an uninformative way as follows. All Wishart parameters $W$ were set to the identity matrix (or 1 in the case of scalar) and the $\nu$ parameters were set to the minimum value $D+1$. For the Gaussian distribution, we set the mean to zero vector and the covariance to identity. Finally, we adopted a reward model that consists of the Euclidean distance to the goal and a quadratic cost term for the actions.

The RL progress for both our method and VICES are shown in Fig. \ref{fig:blocks_all_tasks}. We see that our method achieves better sample efficiency than VICES in all cases. Note that VICES required at least 30 rollouts per iteration and hence ended up having double the number of total rollouts. Task1 for VICES showed slight decline after initial progress; recall that insertion clearance for Task1 is the tightest so even a slight change can make a difference. Task3 was not successful for both our method (even for $K=16$) and VICES; we believe this task can only be solved with reward shaping. See \cite{buchli2011learning} for an example of shaping reward functions with multiple waypoints. We avoided this approach because our focus is on discovering close distance interaction behavior rather than complex motion profiles. 

Does our method learn variable impedance at all? From the plots of the combined stiffness (Fig. \ref{fig:blocks_task2_energy_comp}b) and damping (Fig. \ref{fig:blocks_task2_energy_comp}c) matrices, along a successful rollout in the final iteration for Task2, we see that both stiffness and damping matrices have larger eigenvalues (higher impedance) at the beginning of the trajectory and gets smaller (lower impedance) at the vicinity and the interior of the slot. This is exactly what one would expect: higher impedance for free motion and lower impedance for contact motion. Note that the trajectory indicates free motion up until it makes contact followed by a smooth insertion. The first contact is visible as a small blip right before the insertion.

To test that stability is indeed maintained throughout the training, we plot the Lyapunov function $V(s,\dot{s})$ of iMOGIC, which shares parameters with the policy, for one rollout each from the first and last iterations (Fig. \ref{fig:blocks_task2_energy_comp}a). We see that both plots are monotonically decreasing, indicating the main stability guaranteeing property $\dot{V}<0$. Note that the plot of the first iteration case did not converge to zero indicating that it did not succeed. This is an example of the case where stability is preserved even when a passive environment is preventing convergence to the equilibrium point. In Fig. \ref{fig:blocks_task2_energy_comp}d-e, we see how RL has reshaped an initial Lyapunov function to one that is rich enough to succeed in the task--without compromising stability.

\subsection{Peg-in-hole in Simulation}\label{sec:result_yumisim}
\begin{figure}[t]
    \centering
    \includegraphics[width=.45\textwidth]{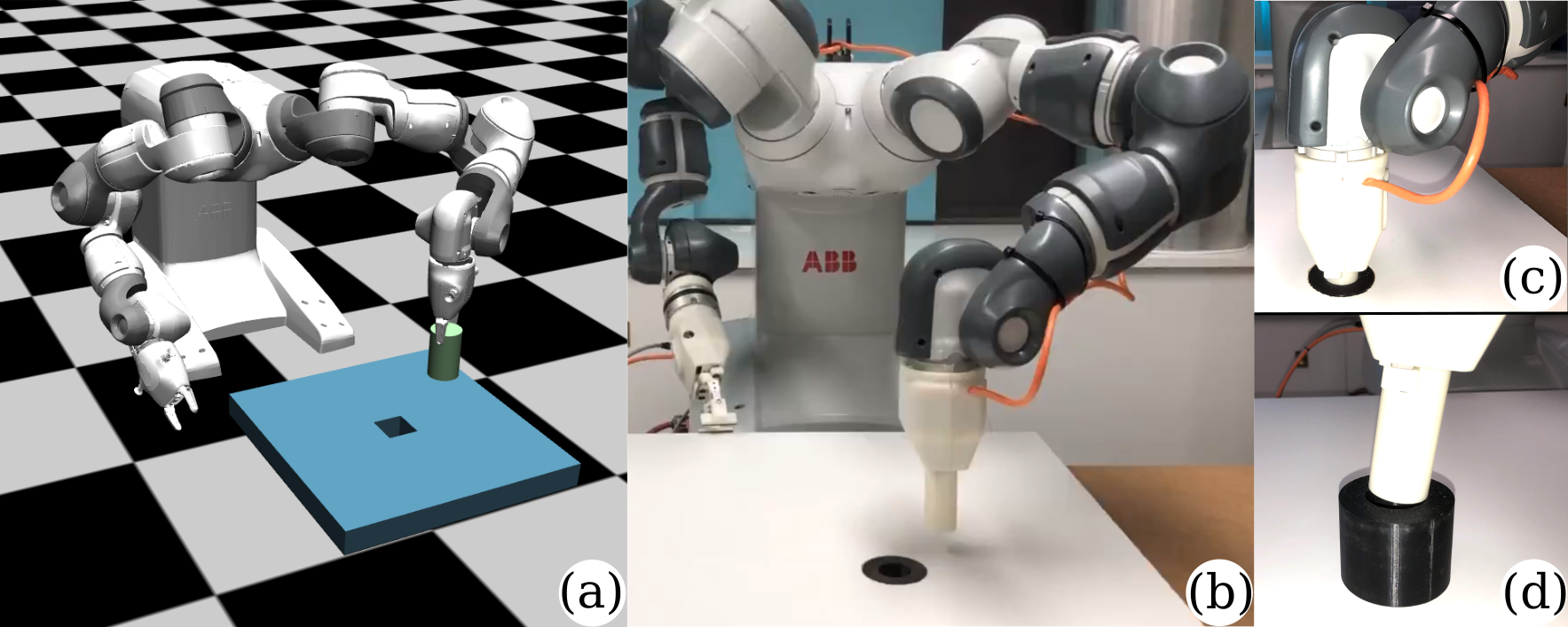}
    \caption{\textbf{Peg-in-hole task} \textbf{(a)} MuJoCo environment: cylindrical peg (24 mm wide, 50 mm long), square hole (25 mm wide, 30 mm deep), and execution time of 2 s \textbf{(b-d)} Real-world environment: cylindrical peg (27 mm wide, 50 mm long), cylindrical hole (27.5 mm wide, 40 mm deep), and 5 s of execution time. \textbf{(c)} Successful insertion position. An insertion depth of 20 mm is considered successful. \textbf{(d)} 3D printed cylindrical peg and hole.}
    \label{fig:yumi_exp_setup}
\end{figure}

In this experiment we scale up to a 7-DOF manipulator arm that is expected to insert a cylindrical peg into a square hole (Fig. \ref{fig:yumi_exp_setup}a). The peg-in-hole task has been historically considered as the benchmark problem for contact-rich manipulation~\cite{vgullapalli1992learning,nuttin1997learning}. To ensure that the policy learns to exploit and also comply with environmental constraints, the initial position is chosen to be laterally away from the hole. The policy is implemented in the operational space including both translation and rotation according to Sec. 9.2.2. of \cite{siciliano2010robotics}, which uses the goal frame as the reference frame and adopts $XYZ$ Euler angle representation. We restrict the Euler angle ranges to $\pm\pi/2$ with respect to the fixed goal frame to avoid representational singularities. This range is not practically limiting for most applications and we assume the rotation to remain in this range once it is initialized appropriately. Although more sophisticated representations such as rotation matrix and quaternions are interesting, i-MOGIC only supports Euler angles. As mentioned earlier, we fix the population parameters as $N_s=15$ and $N_e=3$. Other settings include $K=2$ and $\beta=10$. We adopt the reward model suggested in \cite{levine2015learning}. 

Unlike the previous experiment, where we did not initialize $q(\theta|\Phi)$ informatively, we now propose the following principled approach for an informative initialization. The base spring-damper distribution parameters $W_{\boldsymbol{S}^0},W_{\boldsymbol{D}^0}$ are initialized as, but not constrained to, diagonal matrices such that a suitable critically damped pair would bring the manipulator to a close vicinity of the goal in the desired time. For free motion such a value can be calculated analytically, but in the contact-rich case, an experimental approach is best. The parameters $\nu_{\boldsymbol{S}^0},\nu_{\boldsymbol{D}^0}$ indicate the level of confidence for $W_{\boldsymbol{S}^0},W_{\boldsymbol{D}^0}$ and we used an initial value of $30$. For the remaining spring-damper components, the parameter $W_{\boldsymbol{S}^k}$ was initialized to a fraction of the base parameter ($W_{\boldsymbol{S}^k}=\frac{1}{4}W_{\boldsymbol{S}^0}$) and $W_{\boldsymbol{D}^k}$ set to the corresponding critically damped value. For each of the local attractor points, its Gaussian distribution mean $\mu_{s ^k}$ is initialized to the zero vector (goal point) and the covariance $\Sigma_{s ^k}$ is initialized with a diagonal matrix such that the initial position of the manipulator is one unit of Mahalanobis distance away from the goal. Initializing $q(\theta|\Phi)$ with an informed $\Phi^{init}$, such as above, is crucial for ES based methods (see Sec. \ref{sec:disc}).

\begin{figure}[t]
     \centering
     \begin{subfigure}[b]{0.49\textwidth}
         \centering
         \includegraphics[width=\textwidth]{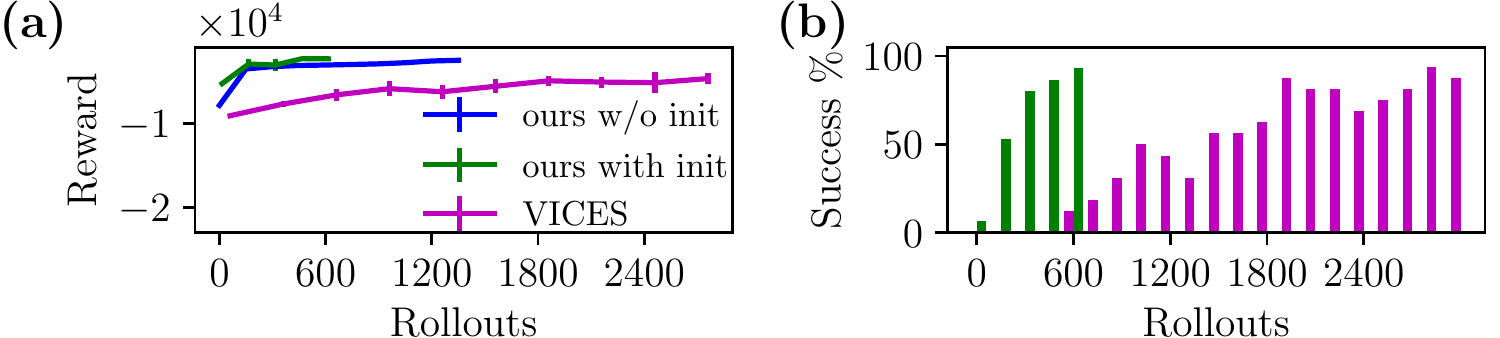}
     \end{subfigure} \\
     \begin{subfigure}[b]{0.49\textwidth}
         \centering
         \includegraphics[width=\textwidth]{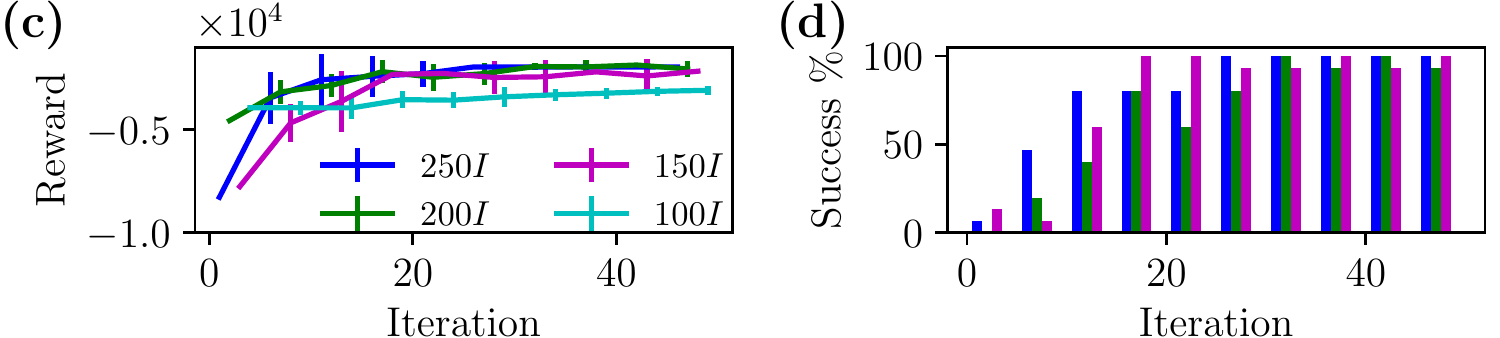}
     \end{subfigure} \\
      \begin{subfigure}[b]{0.49\textwidth}
         \centering
         \includegraphics[width=\textwidth]{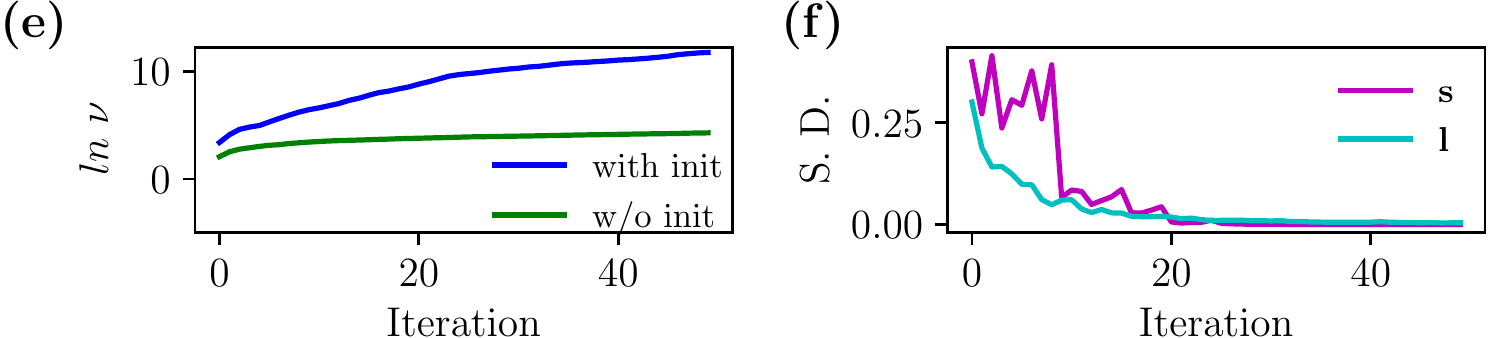}
     \end{subfigure}
     \caption{\textbf{RL on simulated peg-in-hole} \textbf{(a-b)} RL progress comparison. Success rate is among all rollouts per iteration. \textbf{(c-d)} Robustness to variation in initial value of $W_{S^0}$. \textbf{(c)} Growth curve of the $\nu$ parameter \textbf{(d)} Average of element-wise standard deviations of $\vec{s}$ and $\vec{l}$ parameters (init case).}
     \label{fig:yumi_sim_results}
\end{figure}
\begin{figure}[t]
    \centering
     \begin{subfigure}[b]{0.155\textwidth}
         \centering
         \includegraphics[width=\textwidth]{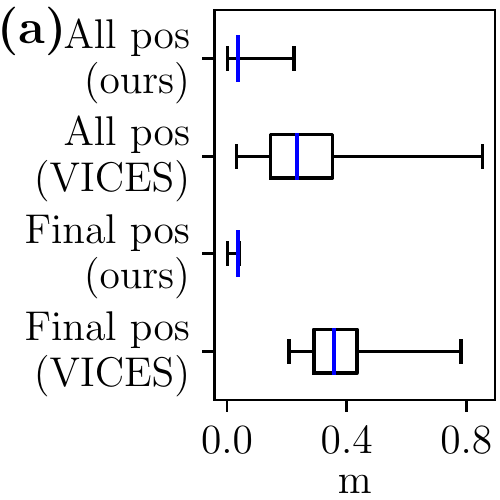}
     \end{subfigure} %
      \begin{subfigure}[b]{0.159\textwidth}
         \centering
         \includegraphics[width=\textwidth,trim=2cm 0cm 1.5cm 0cm, clip]{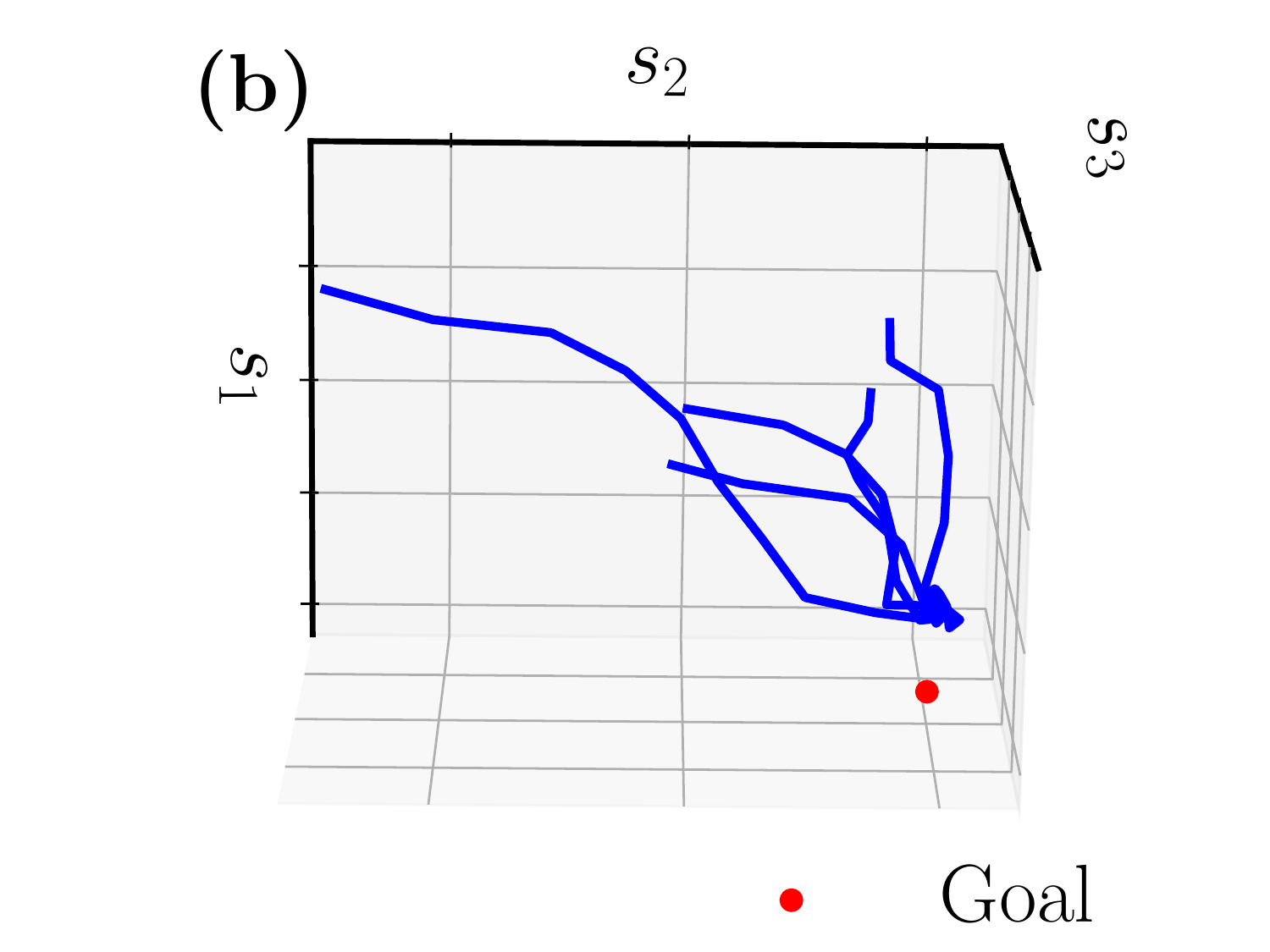}
     \end{subfigure}%
     \begin{subfigure}[b]{0.159\textwidth}
         \centering
         \includegraphics[width=\textwidth,trim=2cm 0cm 1.5cm 0cm, clip]{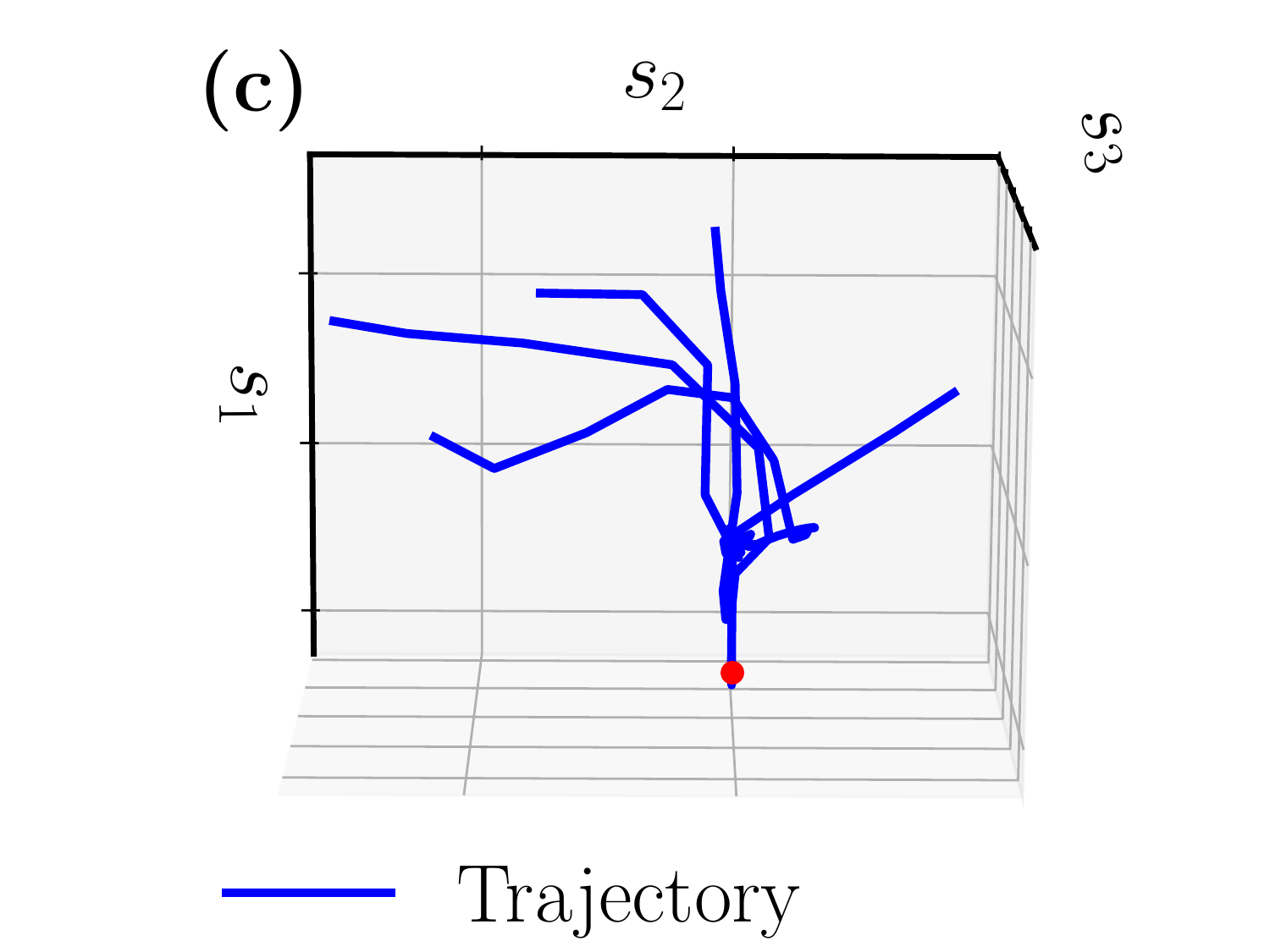}
     \end{subfigure}
    \caption{\textbf{Benefits of stable RL} \textbf{(a)} Distributions of end-effector positions relative to the goal during the first 10 iterations of RL. Whiskers of the box plot include all samples (no outliers). Translation trajectories from five random initial positions show convergence before (b) and after (c) RL.}  
    \label{fig:yumi_mjc_rand_init}
\end{figure}

Fig. \ref{fig:yumi_sim_results}a-b shows examples of our method with and without initialization in addition to VICES. The case with initialization converged in 50 iterations but VICES took about 100 iterations. The case without initialization did not succeed at all. As in the previous experiment, VICES required 30 rollouts per iteration for acceptable performance. Further, we can see in Fig. \ref{fig:yumi_sim_results}e that for the case with initialization, the corresponding $\nu$ parameter grows to a high value ($ln\ \nu\approx 12$) indicating a corresponding reduction in entropy and consequently variance. In Fig. \ref{fig:yumi_sim_results}f we compare the average of element-wise standard deviations of the vectors $\vec{s}$ and $\vec{l}$, which contains parameters $s^k$ and $l^k$, respectively. Note that $\vec{s}$ is modeled by a Gaussian distribution that is updated by MLE and $\vec{l}$ is modeled by a Wishart distribution that is updated by our novel update rule. Remarkably, both cases appear to converge to zero at a similar rate. Fig. \ref{fig:yumi_sim_results}c-d show the result of variations in the initial value of parameter $W_{S^0}$ (rotation component is uninformatively set to I) around the original value of 200I, which was obtained through our initialization scheme. It indicates sufficient robustness with only one case (100I) failing. Note that all other initial parameters are derived from $W_{S^0}$. Since VICES cannot not be informatively initialized, we cannot draw conclusions regarding sample efficiency in this case.

It is also interesting to see the stability property in action. Fig. \ref{fig:yumi_mjc_rand_init}a reveals the real benefit of stability guarantee in RL. We focus on the first 10 iterations of RL because it is during the early stages that stability pays off the most. We see that a lack of stability guarantee can cause potentially unsafe and unpredictable behavior as the robot end-effector covers large distances in the working range of the robot during exploration. It is also able to reach almost any final position. In stark contrast to VICES, our method restricts the robot to a significantly smaller region of the workspace and also forces it to consistently reach the nearest possible final position to the goal that is allowed by the environment. In Fig. \ref{fig:yumi_mjc_rand_init}b-c, we plot some trajectories for particular samples of policy parameters in the first (\ref{fig:yumi_mjc_rand_init}b) and last (\ref{fig:yumi_mjc_rand_init}c) iterations. Individual trajectories correspond to randomly chosen initial positions. Even without any learning (first iteration) the trajectories converge towards the goal position (\ref{fig:yumi_mjc_rand_init}b). Similar behavior is evident for the fully learned final iteration also (\ref{fig:yumi_mjc_rand_init}c). In fact, for the latter case, one of the rollouts even succeeded, indicating a potential for generalizing to initial positions.

\subsection{Peg-in-hole with Real Robot}\label{sec:result_yumireal}
In this experiment we bring the previous simulated experiment of our method into the real world with some minor differences in the geometry and the initial position (Fig. \ref{fig:yumi_exp_setup}b-d). Note that the insertion task is even more challenging with a clearance of only 0.5 mm. We use the same hyperparameter and parameter initialization values as before. A stopping condition for the RL process is introduced: 10 success out of 15 trials--provided that the reward has plateaued--is adopted.
\begin{figure}[t]
    \centering
    \includegraphics[width=.49\textwidth]{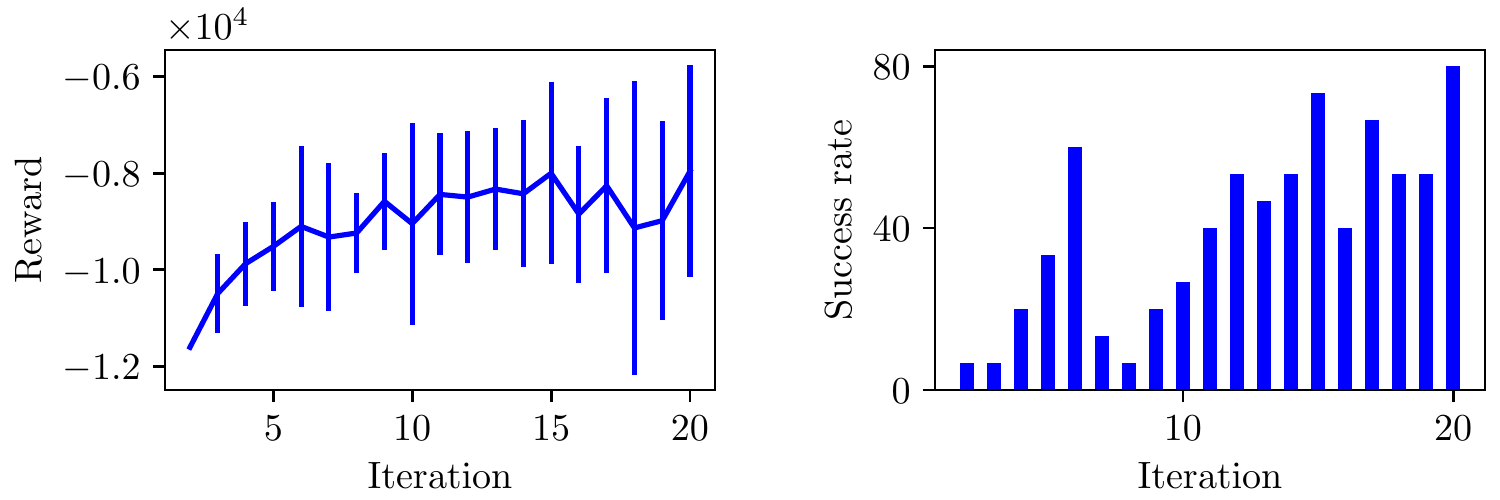}
    \caption{\textbf{RL progress for peg-in-hole with real robot.} Success rate is \% success among all rollouts per iteration.}  
    \label{fig:yumi_rl_progress}
\end{figure}

The stopping condition was satisfied in iteration 20 with 11 successes out of 15 trials. The overall progress of RL can be seen in Fig. \ref{fig:yumi_rl_progress}. Interestingly, due to good initialization, there were instances of insertions very early on; but, as it is evident from the success rate plot, it was also unreliable. The parameter $\nu$ reached a value of about 2000 at iteration 20. We recommend a final target value for $\nu$ in the interval $[1\times10^3, 10\times10^3]$ as a guiding principle for tuning the learning rate $\beta$. Throughout the RL process, the motion exhibited interaction behaviour that relied on environmental constraints for task execution. For example, the spike in success rate at iteration 5 corresponded to the behavior in which the peg slid (on the horizontal surface) all the way to the hole. Later iterations learned a faster strategy to target the opening of the hole directly which also resulted in complex interactions. Stable behavior, such as represented in Fig. \ref{fig:yumi_mjc_rand_init}, was also evident throughout the RL process.

\section{DISCUSSION}\label{sec:disc}
Our results show that it is not only possible to achieve \textit{all-the-time-stability} for RL in the context of contact-rich manipulation but that it also has important benefits. The most important benefit is that stability guarantee can lead to safe and predictable behavior at all stages of RL as evidenced by Fig. \ref{fig:yumi_mjc_rand_init}a. It may be possible to achieve such a behaviour for regular RL methods through prior simulation training or fine tuning of hyperparameters, but these strategies are only best-effort and do not offer any guarantees. In our case, such behaviors are a fundamental property that does not depend on any tuning. It is also worth pointing out that the very definition of stability ensures that the goal position of an episodic task is already built-in to the policy, unlike the case in deep RL methods where the goal is built-in to the reward function and it takes many subsequent iterations for that information to be induced into the policy. It is reasonable to assume that this may have been the reason for the better sample efficiency of our method even with uninformative initialization at times.

A natural concern is regarding the model capacity of the specialized policy i-MOGIC. The authors of \cite{khansari2014modeling} do mention that complex motion profiles may not be possible. Our results suggest that learning contact-rich manipulation that focuses more on interaction behavior rather than arbitrary motion profiles is well-suited for the i-MOGIC framework. ANN-based policy can be expected to have better capacity although, as it is usually the case with deep learning, it comes with data inefficiency. One indication of data inefficiency of ANN based VICES was that it required double the number of rollouts in an iteration. Interestingly, we achieved better efficiency for VICES than what was reported in the original work; for example, we achieved $\sim3000$ rollouts in peg-in-hole task while most tasks took $5000$ to $9000$ rollouts (1 rollout = 200 steps) in \cite{martin2019iros}. A unique benefit of deep RL methods is that it can incorporate complex high dimensional state spaces such as image pixels. Stability guaranteed deep RL methods can have the best of both worlds and will be pursued in our future work. 

ES algorithms often require proper initialization of the sampling distribution to succeed. More specifically, the true solution should be likely to be sampled from the very first iteration. If not, the algorithm will converge to any local optimum that is covered in the initial distribution. Thanks to our informative initialization scheme in Sec. \ref{sec:result_yumisim}, which was possible only because of interpretable structures (spring-damper elements) in our policy, it is possible to effectively initialize our method. Note that in the case of our 2d block-insertion task even an uninformative initialization succeeded. Initialization also undoubtedly contributed to high sample efficiency of our method. Our novel update rule in (\ref{eqn:update_law}) has an additional benefit for real-world RL that imposes constraints on values for $N_s$ and $N_e$. Notice that in Fig. \ref{fig:yumi_sim_results}f, MLE with few samples of $N_e$ ($3$ in our case) has a tendency to collapse the sampling distribution rapidly and thus contribute to local optima. On the other hand, our method offers a fine control on the process through the learning rate $\beta$.

\section{CONCLUSION}
\label{sct:conclusions}
In this paper, we set the ambitious goal of attaining stability-guaranteed reinforcement learning for contact-rich manipulation tasks. To include stable exploration into our notion of stability, we introduced the term \textit{all-the-time-stability}. We built upon a previously proposed stable variable impedance controller and developed a novel model-free policy search algorithm that is inspired by Cross-Entropy Method. We formulated the sampling distribution such that it inherently satisfied stability conditions and then derived an update law for its improvement. The experimental results are significant because \textit{all-the-time-stability} was achieved and its benefits demonstrated on the benchmark task of peg-in-hole without sacrificing sample efficiency.

\bibliographystyle{IEEEtran}
\bibliography{References/rl,References/rl_skill,References/rl_skill_compliant,References/other,References/imitation_learning,References/control_opt_robotics,References/model_learning,References/ml,References/IEEEabrv,References/for_this_doc}

\end{document}